\pdfoutput=1

\documentclass[11pt]{article}

\usepackage[final]{acl}

\usepackage{times}
\usepackage{latexsym}

\usepackage[T1]{fontenc}

\usepackage[utf8]{inputenc}

\newcommand{\benchmark}{NoisyToolBench\xspace}
\newcommand{\method}{AwN\xspace}
\newcommand{\evaluator}{ToolEvaluator\xspace}
\usepackage{adjustbox}
\usepackage{arydshln}
\usepackage{enumitem}
\usepackage{multirow}
\usepackage{tcolorbox}
\usepackage{colortbl}
\usepackage{booktabs}
\usepackage{subfig}

\usepackage{subfloat}

\usepackage{makecell}

\usepackage{xspace}
\usepackage{float}
\usepackage{tipa}

\usepackage[T1]{fontenc}
\usepackage{tabularx}
\usepackage{threeparttable}
\usepackage{pifont}

\usepackage{microtype}

\usepackage{inconsolata}

\usepackage{graphicx}

\title{Learning to Ask: When LLM Agents Meet Unclear Instruction}

\author{Wenxuan Wang$^{1}$ \quad Juluan Shi$^2$\thanks{~~Denote Equal Contribution.} \quad Zixuan Ling$^{2}$$^*$  \quad Yuk-Kit Chan$^{2}$$^*$ \quad Chaozheng Wang$^{2}$  \\ \bf Cheryl Lee$^{2}$ \quad  \bf Youliang Yuan$^{3}$  \quad  \bf Jen-tse Huang $^{4}$\thanks{~~Jen-tse and Wenxiang are corresponding authors.}  \quad \bf Wenxiang Jiao$^{5\dagger}$ \quad \bf Michael R. Lyu$^2$ \\
$^1$Renmin University of China  \quad \quad $^2$The Chinese University of Hong Kong\\
$^3$The Chinese University of Hong Kong, Shenzhen \\
$^4$Johns Hopkins University \quad $^5$Xiaohongshu Inc. \\
$^1$\texttt{wangwenxuan@ruc.edu.cn} \quad \quad $^4$\texttt{jhuan236@jh.edu} \quad \quad  $^5$\texttt{wenxiangjiaonju@gmail.com} \\
}

\begin{document}
\maketitle
\begin{abstract}
Equipped with the capability to call functions, modern LLM agents can leverage external tools for addressing a range of tasks unattainable through language skills alone. 
However, the effective execution of these tools relies heavily not just on the advanced capabilities of LLM agents but also on precise user instructions, which often cannot be ensured in the real world.
To evaluate the performance of LLM agents tool-use under imperfect instructions, we meticulously examine the real-world instructions queried from users, analyze the error patterns, and build a challenging tool-use benchmark called Noisy ToolBench (\benchmark).
We find that due to the next-token prediction training objective, LLM agents tend to arbitrarily generate the missed argument, which may lead to hallucinations and risks.
To address this issue, we propose a novel framework, Ask-when-Needed (\method), which prompts LLM agents to ask questions to users whenever they encounter obstacles due to unclear instructions. 
Moreover, to reduce the manual labor involved in user-LLM interaction and assess LLM agents' performance in tool utilization from both accuracy and efficiency perspectives, we design an automated evaluation tool named \evaluator.
Our experiments demonstrate that the \method significantly outperforms existing frameworks for tool learning in the \benchmark. 
We release all related code and datasets to support future research\footnote{https://github.com/Mysterchan/learning\_to\_ask}.
\end{abstract}

\section{Introduction}

LLMs have undergone remarkable development since OpenAI introduced ChatGPT-3.5 \cite{bang2023multitask}. This model demonstrates a significant advancement in solving multiple tasks, including code generation \cite{dong2023self, sakib2023extending, feng2023investigating}, machine translation \cite{jiao2023chatgpt, peng2023towards}, even game playing \cite{wu2024smartplay}.  However, despite their impressive capabilities, LLMs often struggle with complex computations and delivering accurate, timely information~\cite{qu2024tool}. Tool learning emerges as a promising solution to mitigate these limitations of LLMs by enabling dynamic interaction with external tools~\cite{schick2024toolformer}.

The incorporation of tool usage capabilities marks a pivotal step towards enhancing the intelligence of LLMs, pushing them closer to exhibiting human-like intelligence. The integration of tool usage allows LLMs to perform a broader array of complex and varied tasks. For example, LLMs can perform complex calculations
using a calculator tool, access real-time weather updates through weather APIs, and execute programming code via interpreters \cite{qin2023tool, schick2024toolformer, mialon2023augmented, yang2023foundation}.
Toolformer~\cite{schick2024toolformer} is a pioneering work in empowering language models with self-learning capabilities for tool usage. Then, significant research efforts have been directed toward accessing a wider variety of tools or using multiple tools simultaneously to resolve a single query, such as Gorilla\cite{patil2023gorilla}, RestGPT~\cite{song2023restgpt} and ToolLLM~\cite{qin2023toolllm}.

\begin{figure*}[t]
    \centering
    \subfloat[The execution process of previous frameworks.]{
    \includegraphics[width=0.45\textwidth]{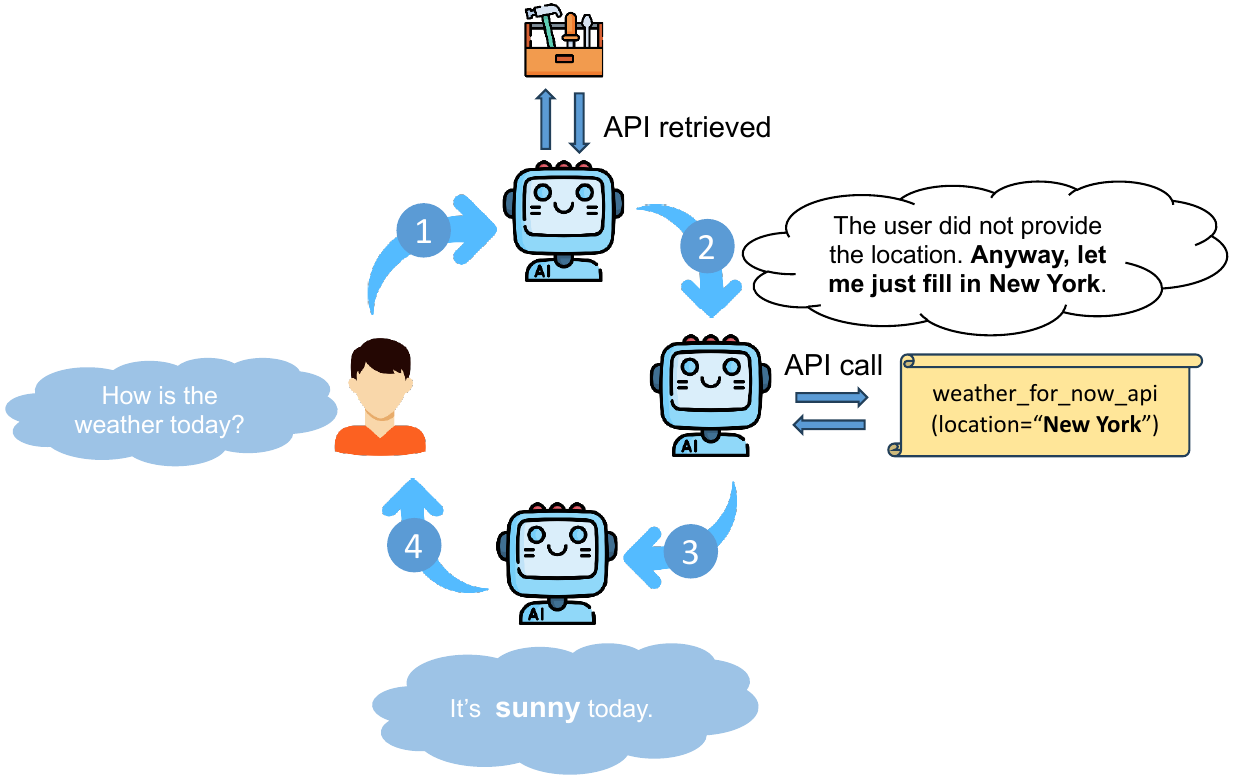}
    \label{fig:issue}
    }
    \subfloat[The execution process of our framework.]{
    \includegraphics[width=0.45\textwidth]{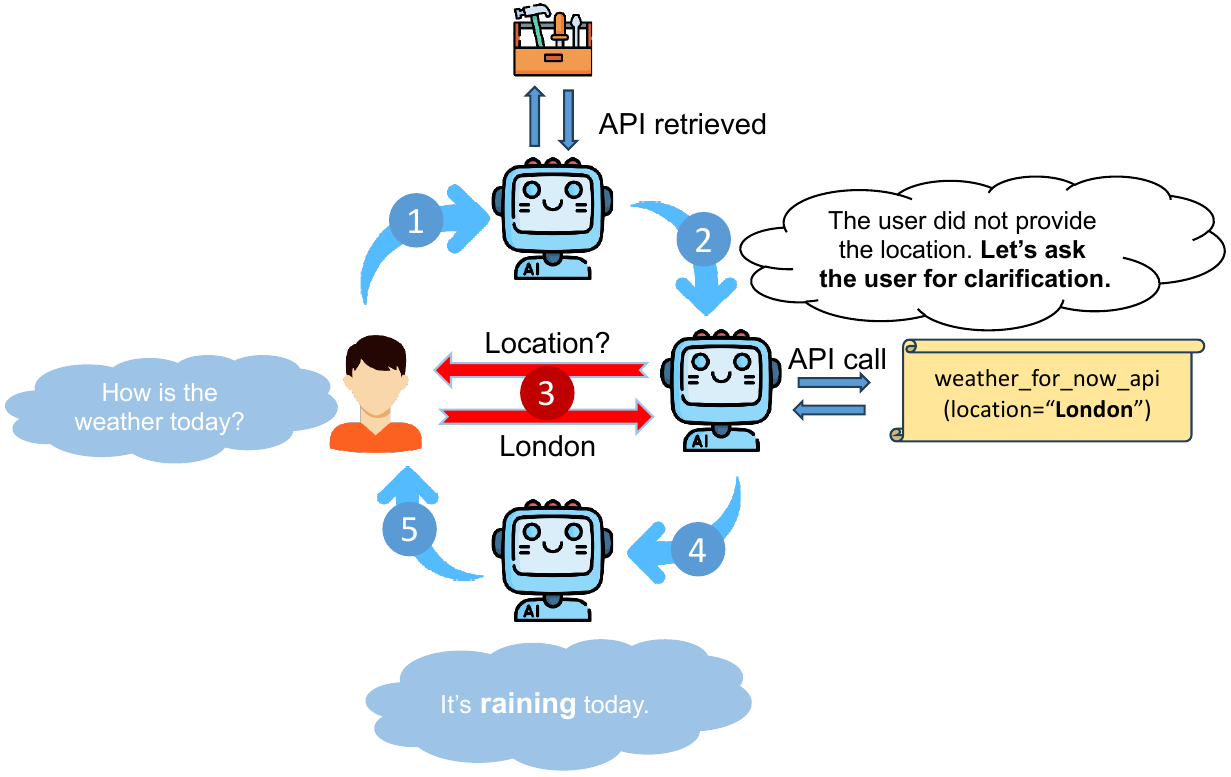}
    \label{fig:solution}
    }
    \caption{The motivating example of our Ask-when-Needed~(AwN) framework.}
    \vspace{-10pt}
\end{figure*}

Despite the significant strides made, existing frameworks and benchmarks often operate under the assumption that user instructions are always explicit and unambiguous, a premise that diverges from real-world scenarios \cite{qin2023tool, song2023restgpt, patil2023gorilla}. 
Due to the feature of API calls, it requires precise user instructions since the arguments for the function call can hardly tolerate ambiguity. We find that due to the next-token prediction training objective, LLMs tend to arbitrarily generate the missed argument, which may lead to hallucinations and risks (as the example shown in Figure~\ref{fig:issue}).
Furthermore, as the tasks assigned to LLMs grow in complexity, multiple and sequential API calls are needed to resolve a single task. This complexity amplifies the challenge, as any error in the sequence of API calls can culminate in an outcome that strays from the user's original intention. 
Hence, LLMs tool-use under unclear instruction is an important but rarely investigated direction.


To address this oversight, we conduct a systematic analysis of actual user instructions, identifying and categorizing potential issues into several key areas. These include instructions lacking essential information, instructions with ambiguous references, instructions containing inaccuracies, and instructions that are unfeasible for LLMs to execute due to the limitations of the tools available. 
Building on this observation, we meticulously design a noisy instruction benchmark, named \benchmark, which is pimarily used for assessing the capability of LLMs to detect ambiguities in user queries and to pose relevant questions for clarification accordingly.
Specifically, \benchmark includes a collection of provided APIs, ambiguous queries, anticipated questions for clarification, and the corresponding responses.

To improve the performance of LLMs tool-use under unclear instructions, we propose a novel framework called Ask-when-Needed (\method). Our key insight is encouraging LLMs to proactively ask questions to seek clarifications from users when uncertainties arise during instruction execution. By facilitating dialogue throughout the process, our method aims to ensure the accurate invocation of functions (See Figure ~\ref{fig:solution})

To evaluate the performance of LLMs tool-use under unclear instruction, we design several innovative metrics from the accuracy and efficiency perspectives. For accuracy, we measure the LLMs' proficiency in asking appropriate clarifying questions, their ability to execute the correct function calls, and their success in delivering final responses that meet the users' needs. For efficiency, we calculate the average number of redundant asked questions and the average number of actions required to complete the instruction. An ideal LLM should achieve higher accuracy with fewer number of queries.
Recognizing the labour-intensive nature of manually communicating with LLMs and verifying all execution results, we also innovatively design an automatic evaluation system, \evaluator, to streamline the whole process. \evaluator\ leverages the advanced problem-solving capabilities of GPT-4o to communicate with LLMs and automatically evaluate the performance of LLMs' tool-using under unclear instructions. Our experiments on 8 LLMs and 2 tool-using frameworks demonstrate that the \method significantly outperforms existing baseline methods. 

The contributions are summarized as follows:
\begin{itemize}[leftmargin=10pt,noitemsep,topsep=0pt]  
\item We conduct a systematic study on real-world user instruction for tool utilization and categorize the prevalent issues into four distinct categories.
    
\item We create and release a novel benchmark, \benchmark, which can be used to evaluate the performance of LLMs' tool-using under imperfect user instruction.

\item We design five evaluation metrics from both accuracy and efficiency perspectives and introduce an automatic evaluation system, \evaluator, that can proxy users to interact and assess LLMs.

\item We introduce a novel framework, named \method method, to prompt LLMs to actively ask questions to request clarifications from users when facing uncertainties. Experimental results show that \method can significantly improve the LLMs' tool-using under unclear instructions.

\end{itemize}

\section{Related Works}

\textbf{Tool Learning for LLMs.} LLMs have recently made significant advancements, with ChatGPT being recognized as a major step towards achieving AGI \cite{wu2023brief, lund2023chatting, jiao2023chatgpt}. However, to progress further towards AGI, it is crucial for LLMs to master the utilization of tools. Toolformer is the first innovative AI model designed to use several specialized tools, such as a web browser, a code interpreter, and a language translator, within a single framework \cite{schick2023toolformer}. The model's ability to seamlessly switch between these tools and apply them contextually represents a significant advancement in AI capabilities. Recent studies like RestGPT~\cite{song2023restgpt} and ToolLLM~\cite{qin2023toolllm}, have connected LLMs with real-life Application Programming Interfaces (APIs), allowing them to sequentially employ multiple external tools to solve user queries. The tool-augmented approach empowers LLMs to use various kinds of tools to do more sophisticated tasks, showcasing an enhanced level of capability compared to pure LLMs. Besides, API-Bank~\cite{li2023api}, ToolAlpaca~\cite{tang2023toolalpaca}, ToolBench~\cite{qin2023toolllm}, ToolQA~\cite{zhuang2023toolqa} and RestBench~\cite{song2023restgpt} are exemplary benchmarks to systematically evaluate the performance of tool-augmented LLMs performance in response to user’s queries. However, current models often ignore the situations in which users might not give exact instructions, which can result in the tools not working properly. Thus, our study aims to tackle this specific challenge by developing a new benchmark specifically for ambiguous instructions. 

\noindent\textbf{Prompting LLMs for Decision Making.} In certain situations, addressing user queries may require more than a single API call. This necessitates the effective division of the overarching task into smaller, more manageable components, which presents a significant challenge. Prior research has focused extensively on enhancing LLMs's ability to effectively plan and execute complex tasks. The 'Chain of Thought' prompting approach facilitates advanced reasoning by introducing intermediate steps in the reasoning process \cite{wei2022chain}. The ReAct methodology improves the integration of reasoning and action, enabling LLMs to take informed actions based on environmental feedback \cite{yao2022react}. Meanwhile, Reflexion is designed to reduce errors in the reasoning process by revisiting and learning from previous mistakes \cite{shinn2023reflexion}. DFSDT expands upon Reflexion, allowing LLMs to evaluate various options and choose the most viable path \cite{qin2023toolllm}. In our work, we innovatively involve users in the process of executing instructions. Our approach, referred to as \method, motivates LLMs to consider the necessity of requesting further information from users during each tool invocation round. This strategy aims at clarifying users' ambiguous instructions to help execute the tasks in alignment with the users' intentions.

\noindent\textbf{Learning to Ask.} Since user queries may not always be clear, and the execution of LLMs may encounter uncertainties and ambiguities, learning to ask questions has emerged as a challenging yet crucial research area~\cite{rao2018learning,kuhn2022clam,andukuri2024star}. For example, some researchers introduce a learning framework that empowers an embodied visual navigation agent to proactively seek assistance\cite{zhang2023good}. Recently, similar ideas have been adopted in the software engineering, leveraging a communicator to enhance the reliability and quality of generated code \cite{wu2023does}. Our work focuses on the tool-learning scenario, which is more sensitive to the user's unclear query. A concurrent study \cite{qian2024tell} also focuses on the reliability of tool-learning systems under unclear instruction. However, they did not systematically examine real-world user behavior, leading to the limited and biased nature of their dataset that doesn't accurately capture user errors. Additionally, Qian's methodology depends significantly on human manual interaction and assessment of LLM performances, which is time-consuming and hard to reproduce.

\section{Noisy ToolBench}
Several tool-learning benchmarks have been introduced to assess LLMs' ability in tool utilization. However, these benchmarks overlook the potential ambiguity in users' instruction, which might hinder LLMs from executing tasks as intended by the user. For instance, as depicted in Figure \ref{fig:issue}, if a user inquires, "How is today's weather" without specifying the location, LLMs cannot accurately activate the APIs to fetch the correct weather information. This scenario underscores the critical role of interaction between users and LLMs in executing instructions accurately. However, previous tool-learning benchmarks only contain perfect user instruction in a one-query-one-execution manner.


To create a realistic benchmark for ambiguous instructions, the initial step involves a systematic examination of the common errors in user instructions that could complicate correct execution by LLMs. Therefore, we first collect real-world user instructions that are problematic. Then, we classify these instructions into various categories based on their characteristics. Lastly, we manually create our dataset, ensuring it reflects the distribution of errors found in the real-world user instructions.

\subsection{User Instruction Analysis}

To analyze the issues in real-world user instruction, we recruit human annotators to write user queries according to the API provided. Firstly, we select 100 APIs from the ToolBench~\cite{qin2023toolllm}, containing real-world RESTful APIs spanning 49 categories, ranging from sports to finance. Secondly, we hire 10 volunteers, who have a Bachelor's degree, are proficient in English, and have experience using LLMs. We provide them with the 100 APIs, and then ask them to write down an instruction to prompt LLMs to call each API, ending up with 1000 user queries. Finally, we manually identify the problematic user queries and categorized them as follows.

\begin{itemize}[leftmargin=10pt,noitemsep,topsep=0pt]
     \item \textbf{Instructions Missing Key Information (IMKI)}: These are user instructions that omit crucial details necessary for the successful execution of a function. An example of IMKI would be, "Set an alarm to wake me up" without providing a specific time. Asking for more information is needed when encountering this issue.
     \item \textbf{Instructions with Multiple References (IMR)}: These user instructions include elements that can be interpreted in several ways, potentially leading to confusion for LLMs in understanding the user's actual intent. For example, an IMR instance is "I want to know the director of the movie 'The Matrix'," where the ambiguity arises because there are multiple versions of 'The Matrix', each possibly having a different director. This issue is similar to IMKI but is more subtle and difficult to detect. Pointing out potential references and asking for clarification are needed when encountering this issue.
     \item \textbf{Instructions with Errors (IwE)}: This category consists of user instructions that contain the necessary information for executing a function, but the information is incorrect. An example of IWE is, "Please help me to log in to my Twitter. My user account is 'abcde@gmail.com' and the password is '123456'," where the user might have provided the wrong account details or password due to typographical errors. Asking for the correct information is needed when encountering this issue.
     \item \textbf{Instructions Beyond Tool Capabilities (IBTC)}: These are user instructions that request actions or answers beyond what LLMs can achieve with the available APIs. In such cases, the existing tool-augmented LLM frameworks might randomly choose an available API, leading to an incorrect function call. This scenario highlights the need for LLMs to recognize their limitations in tool usage. Telling the user that the query is beyond the capabilities and refusing to generate API calls are needed when encountering this issue.
\end{itemize}

 Table~\ref{tab:error_percentage_human} shows the ratio of the four issues, where the most common issue in the instructions is "Instructions Missing Key Information", with a significant 56.0\% of all errors. This issue is a clear indication that users often do not provide adequate information to effectively use the APIs. Additionally, issues such as "Instructions with Errors" and "Instructions Beyond Tool Capabilities" were identified at rates of 17.3\% and 15.3\%, respectively.  

\begin{table}[t!]
\centering
\resizebox{1.0\linewidth}{!}{
\begin{tabular}{lc}
\hline
\textbf{Type of Issue} & \textbf{Ratio} \\ \hline
Instructions Missing Key Information (IMKI)    & 56.0\%                             \\
Instructions with Multiple References (IMR)   & 11.3\%                           \\
Instructions with Errors (IwE)  & 17.3\%                           \\
Instructions Beyond Tool Capabilities (IBTC)       & 15.3\%                           \\ \hline
\end{tabular}}
\caption{Distribution of problematic instructions.}
\label{tab:error_percentage_human}
\vspace{-5pt}
\end{table}
 
    
\subsection{Data Construction}
Our user instruction analysis reveals that there are four kinds of instruction issues that may lead to LLMs' tool utilization failures: Instructions Missing Key Information (IMKI), Instructions with Multiple References (IMR), Instructions with Errors (IwE), and Instructions Beyond Tool Capabilities (IBTC). 
So, we build our benchmark with the four issues by intentionally modifying the problem-free instructions from well-established datasets to problematic ones. We first select 200 data with problem-free instruction from ToolBench and then manually modify the user instructions to make them suffer from the four kinds of instruction issues. Then we annotate the expected questions that LLMs should ask when facing each imperfect user query, which will be used to measure whether LLMs can ask the right questions, as well as the answer to the question, which will be used to proxy the human responses. We conduct a two-round cross-verification to ensure the quality of the annotation. Each data is annotated and verified by different people and any disagreement data will be re-annotated until reach a consensus.
Finally, each data entry in \benchmark has the following five components: the imperfect user query, the available APIs, the questions that LLMs should ideally ask, the answers to these questions, and the expected function calls along with their respective arguments.

\begin{figure}[t]
    \centering
    \includegraphics[width=0.48\textwidth]{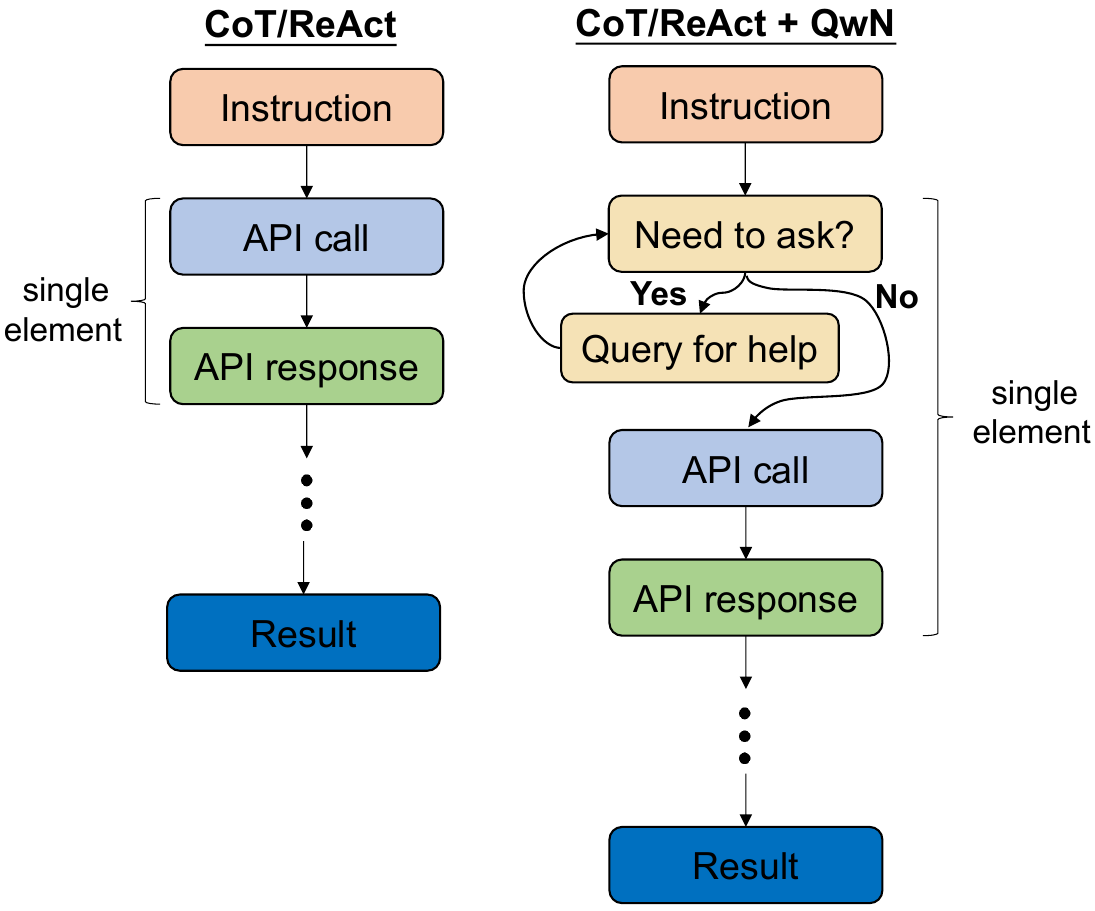}
    \caption{The comparison of our QwN prompting compared with original CoT/ReAct Prompting}
    \label{fig:QwN}
\vspace{-5pt}
\end{figure}

\section{Ask-when-Needed Prompting}
Previous approaches to tool-using often overlooked the importance of user engagement during the reasoning and planning stages. To address this oversight, we introduce a new prompting strategy named Ask-when-Needed (\method). 
The key insight is prompting LLMs to detect the potential flaws in user instructions and proactively seek clarifications by asking questions before generating the API call. 

\method is built upon widely-used tool-using methods, such as CoT and ReAct. As in Figure~\ref{fig:QwN}, we introduce an additional step before the generation of API calls. This step involves presenting all available information to the LLMs, including the user query and API guideline, and prompting them to determine the adequacy and correctness of user instruction. If LLMs identify any missing argument needed for function execution based on the API's requirements, they are encouraged to ask questions to the user for this information. \method prompts LLMs not to generate any API call until obtaining all the necessary information. In other words, only if no further information is needed, they can bypass the query step and directly initiate the API call. We also provide various kinds of specific instructions and demonstration examples for different kinds of instruction issues. 

\vspace{5pt}
\noindent\fbox{
\begin{minipage}{0.95\linewidth}
\small
\texttt{You are AutoGPT, tasked with processing user requests through a variety of APIs you have access to. Sometimes, the information provided by users may be unclear, incomplete, or incorrect. Your main responsibility is to determine if the user’s instructions are sufficiently clear and detailed for effective use of the APIs. Here’s your strategy:\\
1. If user instructions are missing crucial details for the APIs, pose a question to obtain the necessary information.\\
2. If the user’s instructions appear to be incorrect, delve deeper by asking questions to clarify and rectify the details.\\
3. If the user’s request falls outside the capabilities of your current APIs, notify them that you’re unable to meet the request by stating: ”Due to the limitation of the toolset, I cannot solve the question”. \\
...
}
\end{minipage}
}

\begin{figure*}[t]
    \centering
    \includegraphics[width=0.85\textwidth]{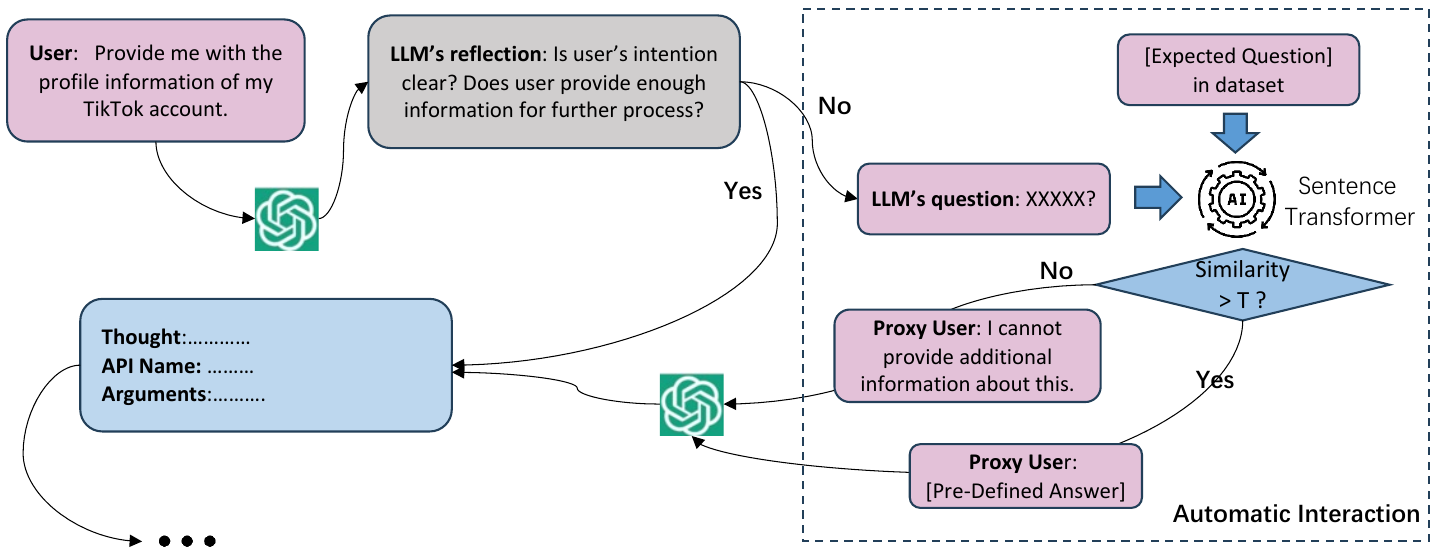}
    \caption{Illustration of the Auto-Interaction module.}
    \label{fig:auto_interaction}
    \vspace{-5pt}
\end{figure*}

\section{Experiments}
In this section, we evaluate the performance of our Ask-when-Needed (\method) prompting technique on the \benchmark dataset. We first introduce the evaluation metrics, where we specify the criteria used to assess the effectiveness of \method. Then, we describe the evaluation pipeline, detailing the step-by-step process employed to measure \method's performance. Lastly, we discuss the main experiments, presenting the results and findings from our comprehensive testing of the \method technique.

\subsection{Evaluation Metrics}
\label{subsec:metrics}
We evaluate the performance of LLMs' tool-using under unclear instructions from two perspectives: accuracy and efficiency. The accuracy assessment aims to measure the LLMs' capability to make correct decisions during the instruction execution phase and to generate accurate final answers. In contrast, the efficiency assessment focuses on the number of redundant decisions made by the LLMs, considering that unnecessary communication could lead to a waste of processing time. Specifically, we design the following five metrics:
\begin{itemize}[leftmargin=10pt,noitemsep,topsep=0pt]   
\item \textbf{Accuracy 1 (A1).} A1 evaluates the capability of LLMs to ask the anticipated questions that pinpoint the ambiguous elements in user instructions. A1 is considered a success if the LLMs manage to ask the correct questions at any point. Conversely, it is deemed a failure if they do not.
    
\item \textbf{Accuracy 2 (A2).} A2 assesses the ability of LLMs to use all available information to invoke the correct API calls. It is deemed a success if the LLMs call all the anticipated APIs with the correct arguments. If they fail to do so, it is considered a failure.

\item \textbf{Accuracy 3 (A3).} A3 measures the ability of LLMs to extract the anticipated information from previous API calls to fulfill the user's instructions. This is achieved and considered a success if the user's instructions are successfully executed. If not, it is regarded as a failure.

\item \textbf{Average Redundant Asked questions (Re).} This metric evaluates the quantity of irrelevant or redundant questions asked by LLMs during the instruction process. Irrelevant questions are those that do not meet the initial expectations of the query, and redundant questions include those that are repetitive or have previously been asked. This metric is crucial for assessing the LLMs' ability to precisely identify the ambiguous aspects of user instructions and to formulate appropriate questions to clarify these uncertainties. The larger the value, the worse the performance.

\item \textbf{Steps.} Steps quantifies the average number of actions required to complete an instruction, including inference generation, asking questions, and conducting API calls. A smaller number indicates fewer unnecessary steps in the instruction execution process, signifying a more efficient and direct approach to accomplishing the task.
\end{itemize}

\begin{table*}[htb]
    \setlength{\tabcolsep}{4pt}
    \centering
    \begin{adjustbox}{width=0.9\textwidth}
    \begin{tabular}{l*{11}{c}}
    \toprule
    \multirow{2}{*}{\bf Model} & \multirow{2}{*}{\bf Framework} & \multicolumn{3}{c}{\bf IMKI} & \multicolumn{3}{c}{\bf IMR} & \multicolumn{3}{c}{\bf IwE} & \multicolumn{1}{c}{\bf IBTC}\\
    \cmidrule(lr){3-5}\cmidrule(lr){6-8}\cmidrule(lr){9-11}\cmidrule(lr){12-12}
    & & A1(\%) & A2(\%) & A3(\%) & A1(\%) & A2(\%) & A3(\%) & A1(\%) & A2(\%) & A3(\%) & A1(\%)\\
    \midrule
    \multirow{4}{*}{gpt-3.5} & CoT & 0.74 & 0.36 & 0.22 & 0.20 & 0.24 & 0.12 & 0.5 & 0.24 & 0.16 & 0.38 \\
     & + \method & 0.74 & 0.44 & 0.24 & 0.86 & 0.46 & 0.20 & 0.74 & 0.48 & 0.28 & 0.48 \\
    \cmidrule(lr){2-12}
     & DFSDT & 0.64 & 0.16 & 0.12 & 0.60 & 0.18 & 0.16  & 0.48 & 0.14 & 0.14 & 0.46  \\
    & + \method  & 0.88 & 0.52 & 0.46 & 0.88 & 0.56 & 0.48 & 0.72 & 0.42 & 0.36 & 0.64  \\
    \midrule
    \multirow{4}{*}{gpt-4} & CoT & 0.74 & 0.48 & 0.32 & 0.72 & 0.52 & 0.36 & 0.52 & 0.26 & 0.24 & 0.34 \\
    & + \method & 0.94 & 0.62 & 0.50 & 0.76 & 0.44 & 0.38 & 0.48 & 0.34 & 0.34 & 0.94 \\
    \cmidrule(lr){2-12}
    & DFSDT & 0.82 & 0.16 & 0.16 & 0.70 & 0.28 & 0.26 & 0.54 & 0.12 & 0.10 & 0.54 \\
    & + \method & 0.80 & 0.56 & 0.48 & 0.80 & 0.50 & 0.44 & 0.52 & 0.38 & 0.36 & 0.94 \\
    \midrule
    \multirow{4}{*}{gpt-4o} & CoT & 0.52 & 0.48 & 0.34 & 0.18 & 0.28 & 0.16 & 0.12 & 0.12 & 0.10 & 0.10 \\
     & + \method & 0.90 & 0.58 & 0.36 & 0.80 & 0.46 & 0.30 & 0.60 & 0.44 & 0.32 & 0.92 \\
    \cmidrule(lr){2-12}
     & DFSDT & 0.58 & 0.20 & 0.18 & 0.26 & 0.18 & 0.16  & 0.18 & 0.06 & 0.04 & 0.08  \\
    & + \method  & 0.88 & 0.60 & 0.46 & 0.90 & 0.52 & 0.36 & 0.64 & 0.46 & 0.38 & 0.94  \\
    \midrule
    \multirow{4}{*}{deepseek-v3} & CoT & 0.44 & 0.40 & 0.20 & 0.24 & 0.28 & 0.24 & 0.10 & 0.14 & 0.14 & 0.30 \\
     & + \method & 0.70 & 0.52 & 0.36 & 0.70 & 0.54 & 0.46 & 0.40 & 0.30 & 0.26 & 0.98 \\
    \cmidrule(lr){2-12}
     & DFSDT & 0.42 & 0.30 & 0.26 & 0.60 & 0.20 & 0.18  & 0.22 & 0.12 & 0.12 & 0.48 \\
    & + \method  & 0.72 & 0.52 & 0.42 & 0.82 & 0.52 & 0.48 & 0.54 & 0.38 & 0.36 & 0.98  \\
        \midrule
    \multirow{4}{*}{gemini-1.5} & CoT & 0.22 & 0.18 & 0.10 & 0.22 & 0.10 & 0.02 & 0.08 & 0.12 & 0.06 & 0.52 \\
     & + \method & 0.86 & 0.40 & 0.18 & 0.74 & 0.24 & 0.08 & 0.58 & 0.28 & 0.22 & 0.68 \\
    \cmidrule(lr){2-12}
     & DFSDT & 0.62 & 0.02 & 0.02 & 0.6 & 0.08 & 0.04  & 0.36 & 0.06 & 0.02 & 0.48  \\
    & + \method  & 0.82 & 0.40 & 0.12 & 0.76 & 0.28 & 0.04 & 0.66 & 0.36 & 0.26 & 0.70  \\
        \midrule
    \multirow{4}{*}{claude-3.5} & CoT & 0.24 & 0.26 & 0.20 & 0.12 & 0.28 & 0.24 & 0.08 & 0.26 & 0.24 & 0.30 \\
     & + \method & 0.54 & 0.5 & 0.5 & 0.32 & 0.30 & 0.24 & 0.34 & 0.34 & 0.26 & 0.88 \\
    \cmidrule(lr){2-12}
     & DFSDT & 0.26 & 0.18 & 0.14 & 0.12 & 0.18 & 0.18  & 0.12 & 0.20 & 0.18 & 0.62  \\
    & + \method  & 0.52 & 0.44 & 0.42 & 0.32 & 0.30 & 0.18 & 0.36 & 0.36 & 0.30 & 0.86  \\
            \midrule
     \multirow{4}{*}{O3-mini} & CoT  & 0.00 & 0.16 & 0.16 & 0.00 & 0.16 & 0.14 & 0.00 & 0.04 & 0.04 & 0.00 \\
     & + \method & 0.78 & 0.54 & 0.52 & 0.76 & 0.54 & 0.48 & 0.58 & 0.42 & 0.32 & 0.40 \\
     \cmidrule(lr){2-12}
     & DFSDT & 0.00 & 0.16 & 0.16 & 0.00 & 0.16 & 0.10 & 0.00 & 0.06 & 0.02 & 0.00 \\
     & + \method & 0.80 & 0.54 & 0.52 & 0.76 & 0.58 & 0.54 & 0.58 & 0.38 & 0.36 & 0.48 \\
             \midrule
    \multirow{4}{*}{DeepSeek-R1} & CoT  & 0.10 & 0.22 & 0.16 & 0.02 & 0.28 & 0.12 & 0.00 & 0.08 & 0.06 & 0.00 \\
    & + \method & 0.80 & 0.52 & 0.34 & 0.60 & 0.54 & 0.22 & 0.50 & 0.34 & 0.22 & 0.84 \\
    \cmidrule(lr){2-12}
    & DFSDT &  0.02 & 0.20 & 0.18 & 0.04 & 0.26 & 0.14 & 0.00 & 0.08 & 0.04 & 0.00\\
    & + \method & 0.86 & 0.62 & 0.42 & 0.62 & 0.42 & 0.26 & 0.54 & 0.38 & 0.22 & 0.76\\
    \bottomrule
    \end{tabular}
    \end{adjustbox}
    \caption{Assessing the accuracy of various LLMs using different prompting methods in our benchmark.}
    \label{tab:autoevl data}
\end{table*}

\subsection{Auto-Evaluation Pipeline}
\label{subsec:pipeline}
To assess how LLMs perform under unclear instructions, interacting with LLMs and making assessments are needed. Previous work employs individuals to interact with and evaluate LLMs throughout the entire evaluation process, which is inefficient and not reproducible. To address this, we design an automated evaluation method named \evaluator to proxy this process. \evaluator can automatically interact with LLMs and assess their performances.
\\[0.6ex]
\noindent\textbf{Auto-Interaction.} \evaluator can proxy the user's communication with LLMs. When LLMs post a question, \evaluator calculates the semantic similarity between the asked question and the expected question by the sentence-transformer~\cite{reimers2019sentence}. If the similarity is higher than a threshold, \evaluator replies with the predefined answer to the LLMs. Otherwise, this question is treated as an irrelevant question and \evaluator replies with a standard reply of "Sorry, I cannot provide additional information about this.". 
This approach streamlines the evaluation process by reducing the need for human interaction with LLMs, as illustrated in Figure \ref{fig:auto_interaction}.
\\[0.6ex]
\noindent\textbf{Auto-Assessment.} \evaluator can also automatically assess how well LLMs perform under ambiguous instructions according to the five metrics introduced above. 
A1 measures whether LLMs can ask the right question. \evaluator calculates the semantic similarity between the LLMs-asked question and the expected question to asses A1.
A2 measures whether LLMs can conduct correct API calls. Following the previous works~\cite{Yang2023ShadowAT,Chiang2023CanLL,wang2023all, yuan2023gpt}, \evaluator adopts GPT-4o as a judge to identify whether the generated API calls are the same as the expected API calls. 
A3 measures whether LLMs can correctly generate the final answer. \evaluator adopts GPT-4o as a judge to identify whether the final answer aligns with the user's intent.
For measuring the efficiency, \evaluator counts the number of generated irrelevant questions as Re and counts the total number of actions during the process as Steps. 

\begin{table}[htb]
    \setlength{\tabcolsep}{2pt}
    \centering
    \resizebox{1.0\linewidth}{!}{
    \begin{tabular}{l*{9}{c}}
    \toprule
    \multirow{2}{*}{\bf Model} & \multirow{2}{*}{\bf FrWork} & \multicolumn{2}{c}{\bf IMKI} & \multicolumn{2}{c}{\bf IMR} & \multicolumn{2}{c}{\bf IwE} & \multicolumn{2}{c}{\bf IBTC}\\
    \cmidrule(lr){3-4}\cmidrule(lr){5-6}\cmidrule(lr){7-8}\cmidrule(lr){9-10}
    & & Re & Steps & Re & Steps & Re & Steps & Re & Steps  \\
    \midrule
    \multirow{4}{*}{\rotatebox{90}{gpt-3.5}} & CoT & - & 4.46 & - & 4.02 & - & 3.90 & - & 2.10\\
     & + \method & 0.66 & 5.36 & 1.10 & 5.98 & 0.76 & 5.08 & 1.10 & 2.40\\
    \cmidrule(lr){2-10}
  & DFSDT & - & 12.82 & - & 12.80 & - & 13.82 & - & 5.50 \\
    & + \method  & 1.44 & 16.94 & 0.98 & 11.24 & 0.94 & 11.68 & 1.70 & 3.94 \\
    \midrule
    \multirow{4}{*}{\rotatebox{90}{gpt-4}} & CoT & - & 4.00 & - & 3.98 & - & 3.34 & - & 2.04\\
    & + \method & 0.16 & 3.94 & 0.20 & 3.94 & 0.36 & 3.46 & 0.04 & 1.16\\
    \cmidrule(lr){2-10}
    & DFSDT & - & 83.96 & - & 21.04 & - & 22.40 & - & 4.06\\
    & + \method & 0.48 & 9.82 & 0.74 & 13.08 & 0.62 & 9.42 & 0.16 & 2.10\\
    \midrule
    \multirow{4}{*}{\rotatebox{90}{gpt-4o}} & CoT & - & 3.00 & - & 2.98 & - & 2.48 & - & 1.28\\
    & + \method & 0.62 & 3.86 & 0.70 & 3.96 & 0.46 & 3.18 & 0.00 & 1.10\\
    \cmidrule(lr){2-10}
    & DFSDT & - & 5.98 & - & 9.58 & - & 5.78 & - & 8.98\\
    & + \method & 0.86 & 6.70 & 1.18 & 7.68 & 0.88 & 8.56 & 0.00 & 1.14\\
    \midrule
    \multirow{4}{*}{\rotatebox{90}{deepseek-v3}} & CoT & - & 4.20 & - & 3.52 & - & 3.12 & - & 1.18\\
    & + \method & 0.20 & 3.88 & 0.06 & 3.60 & 0.04 & 2.92 & 0.02 & 1.10\\
    \cmidrule(lr){2-10}
    & DFSDT & - & 59.08 & - & 41.70 & - & 24.24 & - & 11.64\\
    & + \method & 1.16 & 15.86 & 1.80 & 24.60 & 1.20 & 11.82 & 0.04 & 1.32\\
    \midrule
    \multirow{4}{*}{\rotatebox{90}{gemini-1.5}} & CoT & - & 4.00 & - & 4.12 & - & 2.86 & - & 4.52\\
    & + \method & 0.42 & 6.44 & 0.68 & 6.36 & 0.48 & 4.54 & 0.46 & 1.46\\
    \cmidrule(lr){2-10}
    & DFSDT & - & 750.80 & - & 685.00 & - & 725.14 & - & 559.78\\
    & + \method & 5.34 & 445.16 & 9.08 & 532.56 & 1.94 & 411.18 & 0.46 & 1.46\\
    \midrule
    \multirow{4}{*}{\rotatebox{90}{claude-3.5}} & CoT & - & 2.64 & - & 3.40 & - & 3.04 & - & 1.90\\
    & + \method & 0.18 & 3.74 & 0.33 & 1.03 & 0.36 & 3.76 & 0.10 & 1.68\\
    \cmidrule(lr){2-10}
    & DFSDT & - & 3.34 & - & 9.64 & - & 5.98 & - & 4.26\\
    & + \method & 0.76 & 6.74 & 0.80 & 17.46 & 1.04 & 13.08 & 0.14 & 2.76\\
    \midrule
    \multirow{4}{*}{\rotatebox{90}{O3-mini}} & CoT & - & 1.88 & - & 3.08 & - & 2.72 & - & 1.08\\
    & + \method & 0.42 & 3.54 & 0.52 & 3.82 & 0.44 & 3.66 & 0.30 & 1.54\\
     \cmidrule(lr){2-10}
    & DFSDT & - &  2.04 & - & 4.36 & - & 2.88 & - & 1.04\\
     & + \method &  0.44 & 3.62 & 0.40 & 4.10 & 0.48 & 4.16 & 0.24 & 1.54\\
      \midrule
    \multirow{4}{*}{\rotatebox{90}{DeepSeek-R1}} & CoT & - & 1.30 & - & 1.22 & - & 1.30 & - & 1.00 \\
    & + \method & 0.16 & 2.50 & 0.20 & 1.94 & 0.12 & 2.00 & 0.02 & 1.02 \\
     \cmidrule(lr){2-10}
    & DFSDT & - & 1.26 & - & 2.16 & - & 2.16 & - & 1.16 \\
     & + \method & 0.22 & 2.54 & 0.36 & 4.08 & 0.4 & 3.66 & 0.00 & 1.02 \\
    \bottomrule
    \end{tabular}
    }
    \caption{Assessing the efficiency of various LLMs using different prompting methods in our benchmark.}
    \label{tab:step data}
    \vspace{-5pt}
\end{table}

\subsection{The Effectiveness of \evaluator}

Since \evaluator is an automatic evaluation method, the evaluation can be inaccurate due to the imperfect nature of AI techniques, such as sentence transformer or GPT-4o as the judge. In this section, we conduct a human annotation to validate the effectiveness of \evaluator. Specifically, we randomly select 50 cases and ask annotators to assess the accuracy and efficiency, according to the evaluation metrics mentioned above. Then we compare the assessment results from \evaluator and human annotators. \evaluator achieves 90\% accuracy, indicating its effectiveness.

\subsection{Experimental Setup}
We evaluated the performance of \method against two baseline methods, chain-of-thought (CoT) \cite{wei2022chain} and depth-first search-based decision tree (DFSDT) \cite{qin2023toolllm}, which are two widely-used tool-learning methods. All the experiments are conducted with several LLMs as engines, gpt-3.5-turbo-0125, gpt-4-turbo-2024-04-09, gpt-4o-2024-11-20, deepseek-v3, gemini-1.5-flash-latest and claude-3-5-haiku-20241022, using the default setting. 
Since an ideal reaction under Instructions Beyond Tool Capabilities (IBTC) is telling the user that the query is beyond the capabilities and refusing to generate API calls, its performance in A2 and A3 are measured neither.

\subsection{Main Result}

We evaluate the performance of \method as well as the baseline methods on our \benchmark dataset. The accuracy-related results are shown in Table~\ref{tab:autoevl data} and the efficiency-related results are in Table~\ref{tab:step data}.

\textbf{\method enhances the capability of LLM Agents to ask pertinent questions across different issues.} For example,  as is shown in Table~\ref{tab:autoevl data}, \method improved the A1 scores from 0.52 to 0.90, from 0.18 to 0.80, from 0.12 to 0.60, and from 0.10 to 0.92 for gpt-4o-based CoT as well as from 0.58 to 0.88, from 0.26 to 0.90, from 0.18 to 0.64 and from 0.08 to 0.94 for gpt-4o-based DFSDT.

\textbf{Asking the right question leads to the better generation and execution of API calls.} Besides the significant improvements on A1, \method also achieves considerable performance in generating correct API calls (A2) and returning the expected final answer (A3). For example, \method improved the A2 scores from 0.48 to 0.58, from 0.28 to 0.46, from 0.12 to 0.44 for gpt-4o-based CoT as well as from 0.20 to 0.60, from 0.18 to 0.52, from 0.06 to 0.46 for gpt-4o-based DFSDT.

\textbf{\method can also improve the performance of reasoning models.} Experimental results on two recently released strong reasoning models, OpenAI O3-mini and DeepSeek R1, showing that 1) vanilla models still cannot achieve good performance under unclear user instruction; 2) although vanilla reasoning models do not ask questions, it can sometimes produce the correct function call (A2 > A1), due to their powerful reasoning abilities; 3) \method can significantly improve the performance by prompting LLMs to ask good questions.

\textbf{\method can improve most of the LLM agents without generating excessive unnecessary questions.} As is shown in Table~\ref{tab:step data}, \method only leads to 0.16, 0.20, and 0.36 redundant questions for gpt-4-based-CoT, as well as 0.48, 0.74, and 0.62 redundant questions for gpt-4-based-DFSDT.

However, a few LLM agents tend to ask more irrelevant or redundant questions, as indicated by the higher Re scores in Table~\ref{tab:step data}. For example, in Gemini-1.5-based DFSDT, where the average number of redundant questions is 5.34, 9.08, and 1.94. 
This suggests that while the \method aids in identifying and addressing ambiguities in user instructions, it also leads to a less efficient querying process.

\textbf{\method can reduce the average cost of LLM's tool-using.} The average number of steps measures the cost of LLMs' tool-using. As is shown in Table~\ref{tab:step data}, adopting \method can reduce the number of actions, especially for gpt-4-based DFSDT, deepseek-v3-based-DFSDT and gemini-1.5-based-DFSDT. Although \method can lead to a higher cost for a few LLM agents, such as claude-3.5, considering the significant performance improvements achieved, the moderate increase in cost is justifiable.

\section{Discussion}
Our study provides several key insights into addressing unclear user instructions in LLM-based tool use.

First, the taxonomy of four ambiguity types—Instructions Missing Key Information (IMKI), Instructions with Multiple References (IMR), Instructions with Errors (IwE), and Instructions Beyond Tool Capabilities (IBTC)—derived from analyzing 1,000 real-world user queries, offers a robust framework for understanding prevalent instruction challenges. Identified through manual annotation by 10 qualified annotators with a two-round cross-verification process, these categories ensure a realistic and controlled evaluation of LLM performance under imperfect instructions. While real-world ambiguities can involve more complex, context-dependent, or cumulative issues in multi-step tasks, our taxonomy, as shown in Table~\ref{tab:error_percentage_human} (e.g., 56.0\% IMKI), captures dominant issues and provides a foundational step for future research into intricate scenarios.

Second, the \method framework, though a simple prompt-engineering approach, demonstrates significant effectiveness in improving LLM tool-use under unclear instructions. By prompting LLMs to proactively seek clarification, \method markedly enhances accuracy across various models (e.g., A1 scores for gpt-4o-based CoT improved from 0.52 to 0.90, Table~\ref{tab:autoevl data}). Its compatibility with diverse LLMs and frameworks like CoT and DFSDT, combined with a 4.2/5 user satisfaction score and 82.5\% resolution rate for unclear queries (Appendix A.3), highlights its practical utility. The simplicity of \method is a strength, offering a scalable baseline for practitioners. Future work could explore fine-tuning LLMs on \benchmark or integrating \method with advanced agent frameworks to further enhance performance under ambiguous instructions.

Finally, \benchmark's focus on unclear instructions is a deliberate design choice to isolate and evaluate LLM performance in challenging scenarios. In real-world settings, LLMs must discern instruction clarity, but our controlled benchmark allows precise assessment of their ability to handle ambiguities. Experiments on clear instructions (Appendix A.1, Table~\ref{tab:clear} and Table~\ref{tab:shortcuts}) confirm that \method maintains performance in such cases, suggesting its robustness across mixed scenarios. Future work could integrate clear and unclear instructions to test LLMs' discernment capabilities, building on \method's foundation to address more dynamic real-world interactions.

\section{Conclusion}
This paper explores how unclear user instructions hinder LLM agents' tool usage by proposing: (1) Noisy ToolBench (\benchmark), a novel benchmark for evaluating LLM performance under ambiguous instructions; (2) Ask-when-Needed (\method), an innovative approach enabling LLMs to request clarification when uncertain; and (3) an automated evaluator (\evaluator) to assess accuracy and efficiency. 
Experimental results show that the \method algorithm significantly improves the performance of LLMs' tool-using under unclear user instructions.


\section*{Limitations}

This paper has two limitations:
\begin{enumerate}[leftmargin=*]
    \item Although \method can improve the performance, there is still a big gap to perfect. We hope that this work can serve as the first stepping stone, inspiring future researchers to delve deeper into this field of study. 
    \item The automatic evaluation process is not 100\% accurate, leading to some potential false negatives and false positives. In the future, more efforts are needed to build a more reliable auto-evaluation method.
\end{enumerate}

\section*{Ethics Statement}

This paper reports a user experience study conducted to evaluate the effectiveness of AwN. All participants were informed in advance that their feedback would be used for research purposes and might appear in this publication. No personal or sensitive data were collected, and all responses were anonymized to protect privacy. Participation was entirely voluntary, and volunteers were selected based on their proficiency with LLMs to ensure informed and meaningful contributions. Our methodology follows responsible research practices, emphasizing transparency, fairness, and respect for participants.

\bibliography{custom}

\appendix

\section{Appendix}
\label{sec:appendix}

\subsection{\method Does Not Affect the Performance on Clear Instructions}

Since \benchmark focuses on evaluating the performance of LLM tool-use agents under unclear user instructions, the data only contains unclear instructions, which can complement existing tool-use evaluation datasets to assess the performance under imperfect user instructions. However, evaluating \method on clear instructions is also important. So we select 200 samples from ToolBench, the user queries of which are clear, and we evaluate the performance of our Awn on this set. As is shown in Table~\ref{tab:clear}, \method can significantly improve the performance on unclear instructions and does not affect the performance under clear instructions. 

To evaluate the generalizability of \method, we tested it on a 200-query subset of ShortcutsBench~\cite{shen2024shortcutsbench}, a benchmark featuring complex real-world API-based tasks with clear instructions. We measured API selection accuracy (correctly choosing the relevant API) and API calling accuracy (executing the API with correct arguments). Table~\ref{tab:shortcuts} shows that \method consistently improves API selection accuracy (e.g., +6.5\% for GPT-4o) and maintains or enhances API calling accuracy (e.g., +4.0\% for Claude) across all models. These results demonstrate that \method not only excels with ambiguous instructions but also generalizes effectively to clear, real-world tool-use scenarios.

\begin{table}[htb]
    \centering
    \resizebox{1.0\linewidth}{!}{
    \begin{tabular}{l*{9}{c}}
    \toprule
    \bf Model  & \bf Clean & \bf Noisy\_IMKI & \bf Noisy\_IMR & \bf Noisy\_IwE \\
    \hline
    GPT-4o & 0.54 & 0.34 & 0.16 & 0.10 \\
    ~~+ \method &  0.51 & 0.36 & 0.30 & 0.32\\
    DeepSeek V3 & 0.64 & 0.20 & 0.24 & 0.14 \\
    ~~+ \method & 0.64 & 0.36 & 0.46 & 0.26 \\
    Claude 3.5 & 0.60 & 0.20 & 0.24 & 0.24 \\
    ~~+ \method &  0.64 & 0.50 & 0.30 & 0.26 \\
    Gemini 1.5  & 0.36 & 0.10 & 0.02 & 0.06 \\
    ~~+ \method & 0.40 & 0.18 & 0.08 & 0.22 \\
    \bottomrule
    \end{tabular}
    }
    \caption{Assessing the performance of \method on both clear and unclear instructions}
    \label{tab:clear}
\end{table}

\begin{table}[htb]
    \centering
    \resizebox{1.0\linewidth}{!}{
    \begin{tabular}{lcc}
    \hline
    \textbf{Method} & \textbf{API Selection Acc.} & \textbf{API Call Acc.} \\
    \hline
    GPT-4o & 86.0 & 45.0 \\
    ~~+ \method & 92.5 & 42.5 \\
    DeepSeek V3 & 89.0 & 53.5 \\
    ~~+ \method & 91.5 & 53.5 \\
    Claude 3.5 & 89.0 & 46.0 \\
    ~~+ \method & 88.5 & 50.0 \\
    Gemini 1.5 & 83.5 & 46.0 \\
    ~~+ \method & 87.0 & 46.0 \\
    \bottomrule
    \end{tabular}
    }
    \caption{Performance of \method on ShortcutsBench with clear instructions (200 queries).}
    \label{tab:shortcuts}
\end{table}

\subsection{Details of the Calculation of Each Metric}

ToolEvaluator evaluated the model performance on A1, Re and Steps by a deterministic, rule-based algorithm that analyzes the responses. A2 and A3 are evaluated by LLM-as-a-judge.

The following are the details of the calculation of each metric.

A1: A1 is computed by analyzing whether the LLMs generate valid clarifying questions under ambiguous instructions in the NoisyToolbench dataset. A binary flag is given for each instruction; it is set to 1 if there is a clarifying question identified by similarity check. Otherwise, it remains 0.

Re: Re is computed by counting the number of redundant questions asked by LLMs. Similar to A1 situations, each clarifying question raised by LLM will be classified into relevant questions for ambiguity resolution or irrelevant questions. Re here is equal to the number of mis-clarifying questions raised by LLMs under each instruction.

Steps: Steps are computed by counting the number of total actions (generate response, ask question and call API) performed by LLM. The algorithm will go through the whole conversation history for each instruction.

A2 and A3: ToolEvaluator adopts LLMs as evaluators. We adopt GPT-4o as the evaluator to score A2 and A3 for objective 1 (whether using the correct API) and objective 2 (whether returning the correct answer). The instruction is:

As an evaluator of tool-augmented language model systems, your responsibility is to assess the models' effectiveness in using APIs to gather the necessary information to fulfill user requests. This involves reviewing the actual API calls made by the models against a given set of required API calls, including their parameters. Your evaluation focuses on two main objectives:

Objective 1: Verify the accuracy of the actual tool calls made by the models against the expected tool calls, including their arguments. A successful outcome means that the models executed all required API calls with expected arguments. Then the score of the Objective 1 should be 1. Otherwise, it is a failure for Objective 1 and the score should be 0. Here are some examples ...

Objective 2: Assess whether the model's final response correctly achieve the user's instruction and check if the final answer was indeed based on the data retrieved from these API calls, as opposed to being generated independently of these tool calls. Success in this objective means the model effectively used the API calls to achieve user's instructions. Conversely, failure suggests the response was not derived from the API data or the user's instruction is not achieved. It's important to note that any thoughts of the LLMs not based on the API response, especially regarding makeup information, are not considered valid answers. Please note that if the model does not provide the final answer, we consider it as a failure case. Here are some examples ...

\subsection{User Experiments}

To show the effectiveness of our method, we follow~\cite{Wang2024UnderstandingUE} to conduct a user experience study with real-world users. Specifically, we recruit 5 volunteers in Section 3.1 back, who have a Bachelor's degree, are proficient in English, and have experience using LLMs. We provided them with 10 APIs, along with the user instructions they designed to call the APIs, as well as the feedback \method generated. Then we ask the following questions: From 1 (very disagree) to 5 (strongly agree), how much do you agree that the system's clarifying questions are relevant and helpful to solving the request? We got an average score of 4.2, showing the effectiveness of \method from the real-world user perspectives.

\end{document}